\documentclass[11pt,a4paper]{article}
\pdfoutput=1
\usepackage[hyperref]{emnlp2018}
\usepackage{times}
\usepackage{latexsym}
\usepackage{amsmath}
\usepackage{booktabs}
\usepackage{url}
\usepackage{pgfplots}
\usepackage{pgfplotstable}
\usepackage{booktabs}
\usepackage{array}
\usepackage{colortbl}
\usepackage{filecontents}
\usetikzlibrary{patterns}
\usetikzlibrary{calc,positioning}
\usepgfplotslibrary{statistics}
\usepgfplotslibrary{groupplots}
\usepackage{tikzsymbols}
\usepackage{multicol}
\usepackage{ulem}
\usepackage{microtype}
\usepackage{paralist}
\usepackage{enumitem}
\setlist{noitemsep}
\usepackage{breqn}
\usepackage{amssymb}
\usepackage{subcaption}

\definecolor{bblue}{HTML}{4F81BD}
\definecolor{rred}{HTML}{C0504D}
\definecolor{ggreen}{HTML}{9BBB59}
\definecolor{ppurple}{HTML}{9F4C7C}
\definecolor{oorange}{HTML}{F08000}

\aclfinalcopy 

\title{Microsoft's Submission to the WMT2018 News Translation Task:\\
How I Learned to Stop Worrying and Love the Data}

\author{Marcin Junczys-Dowmunt \\
Microsoft\\
1 Microsoft Way\\
Redmond, WA 98121, USA}

\date{}

\begin{document}
\maketitle
\begin{abstract}

This paper describes the Microsoft submission to the WMT2018 news translation shared task. We  participated in one language direction -- English-German. Our system follows current best-practice and combines state-of-the-art models with new data filtering (dual conditional cross-entropy filtering) and sentence weighting methods. We trained fairly standard Transformer-big models with an updated version of Edinburgh's training scheme for WMT2017 and experimented with different filtering schemes for Paracrawl. According to automatic metrics (BLEU) we reached the highest score for this subtask with a nearly 2 BLEU point margin over the next strongest system. Based on human evaluation we ranked first among constrained systems. 
We believe this is mostly caused by our data filtering/weighting regime.
\end{abstract}

\section{Introduction}

This paper describes the Microsoft submission to the WMT2018 \cite{bojar-EtAl:2018:WMT1} news translation shared task. We only participated in one language direction -- English-German. Our system follows current best-practice and combines state-of-the-art models with new data filtering and weighting methods.  
According to automatic metrics (BLEU) we reached the highest score for this subtask with a nearly 2 BLEU point margin over the next strongest system. We believe this is mostly caused by our data filtering/weighting regime. 
Based on human evaluation we ranked first among constrained systems. 

Our title references the fact that we built fairly standard models, updating existing baselines for WMT2017 to the new Transformer model  \cite{NIPS2017_7181}, but spent more time on data cleaning and work with Paracrawl. As a side-effect we came up with a new parallel data filtering method which we call dual conditional cross-entropy filtering. 

\section{The Marian toolkit}

For our experiments, we use Marian \cite{marian} an efficient Neural Machine Translation framework written in pure C++ with minimal dependencies.  Microsoft Translator employees are contributing code to Marian. In the evolving eco-system of open-source NMT toolkits, Marian occupies its own niche best characterized by two aspects:
  \begin{itemize}
  \item It is written completely in C++11 and intentionally does not provide Python bindings; model code and meta-algorithms are meant to be implemented in efficient C++ code.
  \item It is self-contained with its own back end, which provides reverse-mode automatic differentiation based on dynamic graphs.
  \end{itemize}

Marian is distributed under the MIT license and available from \url{https://marian-nmt.github.io} or the GitHub repository \url{https://github.com/marian-nmt/marian}.

\section{NMT architectures}

In \newcite{marian}, we prepared a baseline setup for Marian which reproduces the highest scoring NMT system \cite{DBLP:conf/wmt/SennrichBCGHHBW17} in terms of BLEU during the WMT 2017 shared task on English-German news translation \cite{DBLP:conf/wmt/2017}. We further replaced the original RNN-based architecture with  Transformer-style models from \newcite{NIPS2017_7181} corresponding to their ``base'' and ``big'' architectures. In this section, we reuse the recipe and the proposed models as a set of strong baselines. 

\subsection{Deep transition RNN architecture}

  The model architecture in \newcite{DBLP:conf/wmt/SennrichBCGHHBW17} is a sequence-to-sequence model with single-layer RNNs in both, the encoder and decoder. The RNN in the encoder is bi-directional. Depth is achieved by building stacked GRU-blocks resulting in very tall RNN cells for every recurrent step (deep transitions). The encoder consists of four GRU-blocks per cell, the decoder of eight GRU-blocks with an attention mechanism placed between the first and second block. As in \newcite{DBLP:conf/wmt/SennrichBCGHHBW17}, embeddings size is 512, RNN state size is 1024. We use layer-normalization \cite{ba2016layer} and variational drop-out with $p=0.1$ \cite{gal2016theoretically} inside GRU-blocks and attention.

  \subsection{Transformer architectures}
  We very closely follow the architecture described in \newcite{NIPS2017_7181} and their ``base'' and ``big'' models.

  \subsection{Training recipe}

  \begin{table}[t]
  \centering
  \begin{tabular}{lccc}\toprule
  System &	2016 &	2017 & 2018* \\ \midrule
  Deep RNN (x1) & 34.3 & 27.7 & - \\ 
  +Ensemble (x4) & 35.3 & 28.2 & - \\ 
  +R2L Reranking (x4) & 35.9 & 28.7 & - \\ \midrule
  Transformer-base (x1) &	35.6 &	28.8 & 43.2 \\
  +Ensemble (x4) &	36.4 &	29.4 & 44.0 \\ 
  +R2L Reranking (x4) &	36.8 &	29.5 & 44.4 \\ \midrule
  Transformer-big (x1) & 36.6 & 30.0 & 44.2 \\
  +Ensemble (x4) & 37.2 & 30.5 & 45.2 \\ 
  +R2L Reranking (x4) & 37.6 & 30.7 & 45.5 \\ 
  
  \bottomrule
  \end{tabular}
  \caption{BLEU results for our replication of the UEdin WMT17 system for the en-de news translation task.
  We reproduced most steps and replaced the deep RNN model with Transformer models. Asterisk * marks post-submission evaluation. Missing numbers will be provided in final version.}
  \label{wmt-bleu}
  \end{table}

Modeled after the description from \newcite{DBLP:conf/wmt/SennrichBCGHHBW17}, we reuse the example available at \url{https://github.com/marian-nmt/marian-examples} and perform the following steps:

  \begin{itemize}
  \item preprocessing of training data, tokenization, true-casing\footnote{Proprocessing was performed using scripts from Moses \cite{conf/acl/KoehnHBCFBCSMZDBCH07}.}, vocabulary reduction to 36,000 joint BPE subword units \cite{sennrich2016bpe} with a separate tool.\footnote{\url{https://github.com/rsennrich/subword-nmt}}
  \item training of a shallow model for back-translation on parallel WMT17 data;
  \item translation of 10M German monolingual news sentences to English; concatenation of artificial training corpus with original data (times two) to produce new training data;
  \item training of four left-to-right (L2R) deep models (either RNN-based or Transformer-based);
  \item training of four additional deep models with right-to-left (R2L) orientation; \footnote{R2L training, scoring or decoding does not require data processing, right-to-left inversion is built into Marian.}
  \item ensemble-decoding with four L2R models resulting in an n-best list of 12 hypotheses per input sentence;
  \item rescoring of n-best list with four R2L models, all model scores are weighted equally; 
  \item evaluation on newstest-2016 (validation set) and newstest-2017 with sacreBLEU.\footnote{\url{https://github.com/mjpost/sacreBLEU}}
  \end{itemize}
  
At this stage we did not update to WMT2018 parallel or monolingual training data. This might put us at a slight disadvantage, but we could reuse models and back-translated data that was produced earlier. 
  
We train the deep RNN models and Transformer-base models with synchronous Adam on 8 NVIDIA Titan X Pascal GPUs with 12GB RAM for 10 epochs each. The back-translation model is trained with asynchronous Adam on 8 GPUs.
The transformer-big models are trained until convergence on four NVIDIA P40 GPUs with 24GB RAM.  We do not specify a batch size as Marian adjusts the batch based on available memory to maximize speed and memory usage.  This guarantees that a chosen memory budget will not be exceeded during training and uses maximal batch sizes. 

All models use tied embeddings between source, target and output embeddings \cite{press2017using}. Contrary to \newcite{DBLP:conf/wmt/SennrichBCGHHBW17} or \newcite{NIPS2017_7181}, we do not average checkpoints, but maintain a continuously updated exponentially averaged model over the entire training run. Following \newcite{NIPS2017_7181}, the learning rate is set to 0.0003 (0.0002 for Transformer-big) and decayed as the inverse square root of the number of updates after 16,000 updates. When training the Transformer model, a linearly growing learning rate is used during the first 16,000 iterations, starting with 0 until the base learning rate is reached. 

Table \ref{wmt-bleu} contains our results for WMT2017 training data with back-translation. We match results from \newcite{DBLP:conf/wmt/SennrichBCGHHBW17} with our re-implementation of their models (Deep RNN) and outperform them with base and big Transformer versions. Differences between the best Deep RNN model and Transformer-big reach up to 2 BLEU points for the complete system. Ensembling is quite effective, right-to-left reranking seems to be moderately effective for Transformer models. 

\section{Taking advantage of Paracrawl}

This year's shared task included a new, large but somewhat noisy parallel resource: Paracrawl. First experiments with shallow RNN models (chosen for fast experimentation) indicated that adding this data without a rigorous data filtering scheme would lead to catastrophic loss in quality (compare WMT+back-trans and Paracrawl-32M in Table \ref{para-filt}). We therefore experiment with data selection and weighting. 

\subsection{Dual conditional cross-entropy filtering}
\label{dual}

The scoring method introduced in this section is our main contribution to the WMT2018 Shared Task on Parallel Corpus Filtering \cite{parallel-filtering-task:2018:WMT}, details are provided in our  corresponding system submission \cite{mjd-filtering}.

For a sentence pair $(x,y)$ we calculate a score:
\begin{equation}
\begin{aligned}
  & \left| H_{A}(y|x) - H_{B}(x|y) \right| \\
+ & \frac{1}{2} \left( H_{A}(y|x) + H_{B}(x|y) \right) \label{dualscore}
\end{aligned}
\end{equation}
where $A$ and $B$ are translation models trained on the same data but in inverse directions, and $H_M(\cdot|\cdot)$ is the word-normalized conditional cross-entropy of the probability distribution $P_M(\cdot|\cdot)$ for a model $M$:
\begin{equation*}
\begin{aligned}
H_M(y|x) =& -\frac{1}{|y|}\log P_M(y|x) \\
 =& -\frac{1}{|y|} \sum_{t = 1}^{|y|}\log P_M(y_t|y_{<t},x).
\end{aligned}
\end{equation*}

The score (denoted as dual conditional cross-entropy) has two components with different functions: the absolute difference $\left| H_{A}(y|x) - H_{B}(x|y) \right|$ measures the agreement between the two conditional probability distributions, assuming that (word-normalized) translation probabilities of sentence pairs in both directions should be roughly equal. We want disagreement to be low, hence this value should be close to 0. 

However, a translation pair that is judged to be equally improbable by both models will also have a low disagreement score. Therefore we weight the agreement score by the average word-normalized cross-entropy from both models. Improbable sentence pairs will have higher average cross-entropy values. 

This score is also quite similar to the dual learning training criterion from \newcite{NIPS2016_6469} and \newcite{parity}. The dual learning criterion is formulated in terms of joint probabilities, later decomposed into translation model and language model probabilities. In practice, the influence of the language models is strongly scaled down which results in a form more similar to our score. 

While \citeauthor{moore-lewis:2010:Short} filtering requires an in-domain data set and a non-domain-specific data set to create helper models, we require a clean, relative high-quality parallel corpus to train the two dual translation models. We sample 1M sentences from WMT parallel data excluding Paracrawl and train Nematus-style translation models $W_{\mathrm{de}\rightarrow\mathrm{en}}$ and $W_{\mathrm{en}\rightarrow\mathrm{de}}$.

Formula~(\ref{dualscore}) produces only positive values with 0 being the best possible score. We turn it into a partial score with values between 0 and 1 (1 being best) by negating and exponentiating, setting 
$A = W_{\mathrm{de}\rightarrow\mathrm{en}}$ and
$B = W_{\mathrm{en}\rightarrow\mathrm{de}}$:

\begin{dmath*}
\mathrm{adq}(x,y) = \exp(-( \left| H_{A}(y|x) - H_{B}(x|y) \right|\\ 
   + \frac{1}{2} \left( H_{A}(y|x) + H_{B}(x|y) \right)).
\end{dmath*}

We score the entire Paracrawl data with this score and keep the scores. We further assign a value of 1 to all original WMT parallel sentences. That way we have a score for every sentence. 

\subsection{Cross-entropy difference filtering}

We treated cross-entropy filtering proposed by \newcite{moore-lewis:2010:Short} as another score. Cross-entropy filtering or Moore-Lewis filtering uses the quantity
\begin{equation}
\begin{aligned}
H_{I}(x) - H_{N}(x)
\end{aligned}
\end{equation}
where $I$ is an in-domain model, $N$ is a non-domain-specific model and $H_M$ is the word-normalized cross-entropy of a probability distribution $P_M$ defined by a model $M$:
\begin{equation*}
\begin{aligned}
H_M(x) =& -\frac{1}{|x|}\log P_M(x) \\
 =& -\frac{1}{|x|} \sum_{t = 1}^{|x|}\log P_M(x_t|x_{<t}).
\end{aligned}
\end{equation*}
Sentences scored with this method and selected when their score is below a chosen threshold are likely to be more in-domain according to model $I$ and less similar to data used to train $N$ than sentences above that threshold. 

We chose WMT German news data from the years 2015-2017 as our in-domain, clean language model data and sampled 1M sentences to train model $I=W_{\mathrm{en}}$. We sampled 1M sentences from Paracrawl without any previously applied filtering to produce $N=P_{\mathrm{de}}$.

To create a partial score for which the best sentence pairs produce a 1 and the worst at 0, we apply a number of transformations. First, we negate and exponentiate cross-entropy difference arriving at a quotient of perplexities of the target sentence $y$ ($x$ is ignored):
\begin{equation*}
\begin{aligned}
\mathrm{dom}^\prime(x,y) = & \; \exp(-(H_{I}(y) - H_{N}(y))) \\ = & \; \frac{\mathrm{PP}_{N}(y)}{\mathrm{PP}_{I}(y)}.
\end{aligned}
\end{equation*}
This score has the nice intuitive interpretation of how many times sentence $y$ is less perplexing to the in-domain model $W_{\mathrm{de}}$ than to the out-of-domain model $P_{\mathrm{de}}$. 

We further clip the maximum value of the score to 1 (the minimum value is already 0) as:
\begin{equation}
\begin{aligned}
\mathrm{dom}(x,y) = \max(\mathrm{dom}^\prime(x,y), 1).
\end{aligned}
\end{equation}
This seems counterintuitive at first, but is done to avoid that a high monolingual in-domain score strongly overrides bilingual adequacy; we are fine with low in-domain scores penalizing sentence pairs. This is a precision-recall trade-off for adequacy and we prefer precision.

We score the entire parallel data, Paracrawl, back-translated data and previous WMT data with this score. Next we multiply the adequacy and domain-based score to obtain a single score for all parallel data and all Paracrawl data in particular.

\subsection{Data selection}

\begin{table}[t]
\centering
\begin{tabular}{lc}
\toprule
Data & 2016 \\
\midrule
WMT+back-trans. & 32.6 \\ \midrule
+Paracrawl-32M & 30.1 \\ \midrule
+Paracrawl-2M & 33.2 \\
+Paracrawl-4M & 33.5 \\
\bf +Paracrawl-8M & \bf 34.0 \\
+Paracrawl-16M & 31.9 \\
+Paracrawl-24M & 30.3 \\ \midrule
\bf +Paracrawl-8M-weights & \bf 34.2 \\
+Paracrawl-24M-weights & 33.4 \\
\bottomrule
\end{tabular}
\caption{Effects of data cleaning, filtering and weighting on BLEU. Evaluated with default shallow Nematus-style RNN model}\label{para-filt}
\end{table}

Based on the scores produced in the previous section, we sort the new Paracrawl data by decreasing scores from 1 to 0. Next we select the first N sentences from the sorted corpus, add it to WMT and back-translated data and train again a shallow RNN model. In our experiments it seems, that selecting the first 8M out of 32M sentences according to this score leads to the largest gains on WMT2016 test data. A loss of 2.5 BLEU on full WMT+Paracrawl data is turned into a gain of 1.4 BLEU on WMT with selected Paracrawl data (see +Paracrawl-8M in Table~\ref{para-filt}). 

\subsection{Data weighting}

We further experiment with sentence instance weighting, a feature available in Marian. Here we use the computed score for a sentence pair as a multiplier of the per-sentence cross-entropy cost during training. Sentences with high scores will contribute more to the training, sentence with low cost contribute less. Scores are however clipped at 1, so no score can contribute more than it would without weighting. As stated above, sentences from original WMT training data and from back-translation have an adequacy score of 1, so they are only weighted by their domain multiplier. Sentences from Paracrawl are weighted by a product of their adequacy and domain score. We see slight improvements for +Paracrawl-8M-weights over the unweighted version. It also seems that weighting can at least partially eliminate harmful effects from bad data. The 24M variant is far less damaging than the unweighted version. This seems worth to be explored in future work.

\section{Final submission}

\begin{table}[t]
  \centering
  \begin{tabular}{lccc}\toprule
  System &	2016 &	2017 & 2018* \\ \midrule
  Transformer-big (x1) & 38.6 & 31.3 & 46.5 \\
  +Ensemble (x4) & 39.3 & 31.6 & 47.9 \\ 
  +R2L Reranking (x4) & 39.3 & 31.7 & 48.0 \\ \midrule
  \bf +Transformer-LM & \bf 39.6 & \bf 31.9 & \bf 48.3 \\ 
  \bottomrule
  \end{tabular}
  \caption{Best model retrained on WMT and selected Paracrawl data. Sentences are weighted. Asterisk * marks post-submission evaluation.}
  \label{para-bleu}
  \end{table}

We chose the +Paracrawl-8M-weights setting as our training setting for the Transformer-big configuration. Training and model parameters remain the same, we only add 8M Paracrawl sentences and sentence-level scores for all parallel sentences and retrain all models. In Table~\ref{para-bleu}, we see that compared to Table~\ref{wmt-bleu} the Transformer-big model can take even more advantage of the filtered, selected and weighted data than the shallow models we used for development. We gain 1 to 2.5 BLEU points on the different test sets. Right-to-left re-ranking seems to matter less, however these models had not yet fully converge at time of submission.

\subsection{Ensembling with a Transformer-style language model}

We also experiment with shallow-fusion\footnote{We do not like this term, in the end this is just ensembling.} or ensembling with a language model. We train a Transformer-style language model with Marian, following the architecture of the Transformer-big decoder without target-source attention blocks. We observed that this type of model has lower perplexity than LSMT models with similar numbers of parameters. We use 100M German monolingual sentences from 2016-2018 news data and train for two full epochs. 

The resulting language model is ensembled with the left-to-right translation models at decoding time. We determine an optimal weight of 0.4 on a the newstest2016. The other models in the ensemble have a weight of 1. Since scores are summed it is a 4 to 0.4 ratio for translation models versus language model log probabilities. We see that the language model has a small, but consistently positive effect on all test sets of 0.2-0.3 BLEU. 

\section{Results}

\begin{table}[t]
  \centering
  \begin{tabular}{lccc}\toprule
  System &	BLEU \\ \midrule
  \sc\bf Microsoft-Marian & \bf 48.3 \\
  \sc UCAM & 46.6 \\
  \sc NTT & 46.5 \\
  \sc KIT & 46.3 \\
  \sc MMT-production & 46.2 \\
  \sc UEdin & 44.4 \\
  \sc JHU & 43.4 \\
  \bottomrule
  \end{tabular}
  \caption{Automatic BLEU scores from submission page for 7 best submissions. There were 21 submissions in total. }
  \label{10best}
  \end{table}

\begin{table}[t]
  \centering
\begin{tabular}{cccl}\toprule
Rank & Ave. \% & Ave. z & System \\ \midrule
2 \bf   & \bf81.9	& \bf 0.551 &\sc\bf Microsoft-Marian \\
   & 82.3	& 0.537 & \sc UCAM \\
   & 80.2	& 0.491 & \sc NTT \\
   & 79.3	& 0.454 & \sc KIT \\\midrule
8   & 76.7	& 0.377 & \sc JHU \\
   & 76.3	& 0.352 & \sc UEdin \\ \midrule
11 & 71.8	& 0.213 & \sc LMU-nmt \\ \midrule
15 & 36.7	& -0.966 & \sc RWTH-unsup \\ \midrule
16 & 32.6	& -1.122 & \sc LMU-unsup \\ \bottomrule
\end{tabular}
  \caption{Human evaluation of constrained systems. Unconstrained systems have been omitted, see \newcite{bojar-EtAl:2018:WMT1} for full list.}
  \label{human}
  \end{table}

According to the automatically calculated BLEU scores on the WMT submission page, we achieve the highest BLEU score for English-German by a large margin over the next best system. We include the results for the 7 best systems in Table~\ref{10best}. The next best systems are quite tightly packed.
We also rank highest among constrained systems based on human evaluation (Table~\ref{human}).

\section{Conclusions}

It seems strong state-of-the-art models and data hacking are winning combinations. 
Our data filtering method -- developed first for this system -- also proved very effective during the Parallel Corpora Filtering Task and we believe it had a large influence on our current result. 

\bibliography{acl2018}
\bibliographystyle{acl_natbib_nourl}

\end{document}